\def\STS{s$2$s}
\def\STSL{s$2$sL}
\documentclass{article}

\makeatletter
\def\blfootnote{\gdef\@thefnmark{}\@footnotetext}
\makeatother

\usepackage[final]{nips_2017}

\usepackage[utf8]{inputenc} 
\usepackage[T1]{fontenc}    
\usepackage{hyperref}       
\usepackage{url}            
\usepackage{booktabs}       
\usepackage{amsfonts}       
\usepackage{nicefrac}       
\usepackage{microtype}      
\usepackage{graphicx}       

\title{A Novel Approach for Effective Learning in Low Resourced Scenarios\thanks{Presented at NIPS 2017 Machine Learning for Audio Signal Processing (ML4Audio) Workshop,  Dec. 2017}}

\author{
  Sri Harsha Dumpala \\
  TCS Research and Innovation-Mumbai\\
  \texttt{d.harsha@tcs.com} \\
  \And
  Rupayan Chakraborty \\
  TCS Research and Innovation-Mumbai\\
  \texttt{rupayan.chakraborty@tcs.com} \\
  \AND
  Sunil Kumar Kopparapu \\
  TCS Research and Innovation-Mumbai\\
  \texttt{sunilkumar.kopparapu@tcs.com} \\
}

\begin{document}

\maketitle

\begin{abstract}
Deep learning based discriminative methods, being the state-of-the-art machine learning techniques, are ill-suited for learning from lower amounts of data. In this paper, we
propose a novel framework, called simultaneous two sample learning (s$2$sL), to effectively learn the class discriminative characteristics, even from very low amount of data.
In s$2$sL, more than one sample (here, two samples) are simultaneously considered to both, train and test the classifier. We demonstrate our approach for speech/music 
discrimination and emotion classification through experiments. Further, we also show the effectiveness of s$2$sL approach for classification in 
low-resource scenario, and for imbalanced data.
\end{abstract}


\section{Introduction}

Deep neural networks (DNNs), in particular convolutional and recurrent neural networks, with huge architectures have been proven successful in wide range of tasks including
audio processing such as speech to text [1 - 4], emotion recognition [5 - 8], speech/non-speech (e.g., of non-speech include noise, music, etc.,) classification
[9 - 12], etc.

Training these deep architectures require large amount of annotated data, as a result, they cannot be used in low data resource scenarios which is common in speech-based applications [13 - 15].
Apart from collecting large data corpus, annotating the data is also very difficult, and requires manual supervision and efforts. Especially, annotation of speech for tasks
like emotion recognition also suffer from lack of agreement among the annotators [16]. Hence, there is a need to build reliable systems that can work in low resource scenario.

In this work, we propose a novel approach to address the task of classification in low data resource scenarios. Our approach involves simultaneously considering more than
one sample (in this work, two samples are considered) to train the classifier. We call this approach as simultaneous two sample
learning (s$2$sL). The proposed approach is also applicable to low resource data suffering with data imbalance.
The contributions of this paper are:
\begin{itemize}
 \item Representation of the training data, where the feature vectors pertaining to two different samples are simultaneously considered to train the classifier.
 \item Introduce modifications to the multi-layer perceptron (MLP) architecture so that it can be trained using our proposed data representation.
 \item New decision mechanism to classify the test samples.
 \end{itemize}


\begin{figure}[!tbp]
  \centering
  \begin{minipage}[b]{0.47\textwidth}
    \includegraphics[width=\textwidth]{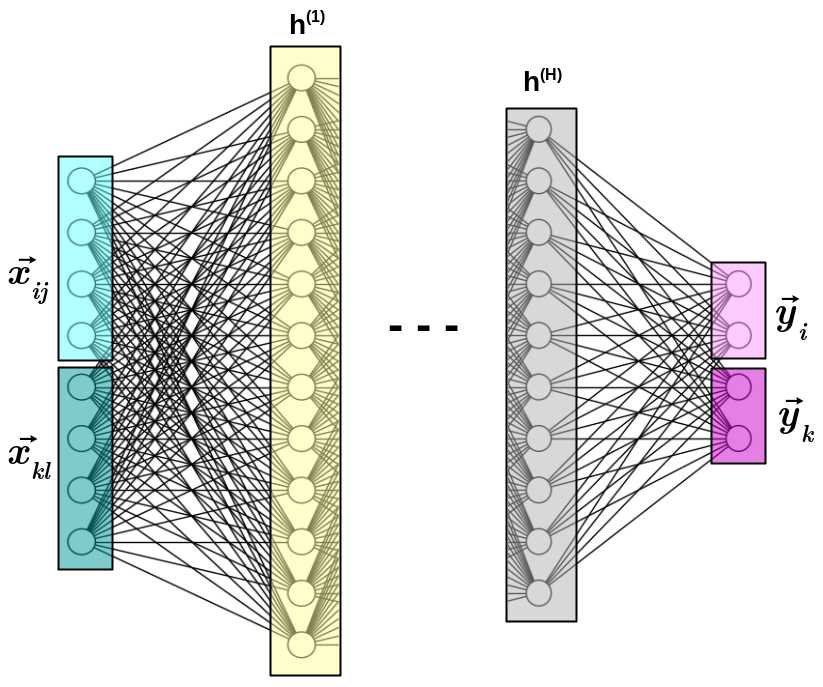}
    \caption{Training of s$2$s-MLP.}
    \label{ProposedMethod}
  \end{minipage}
  \hfill
  \begin{minipage}[b]{0.45\textwidth}
    \includegraphics[width=\textwidth]{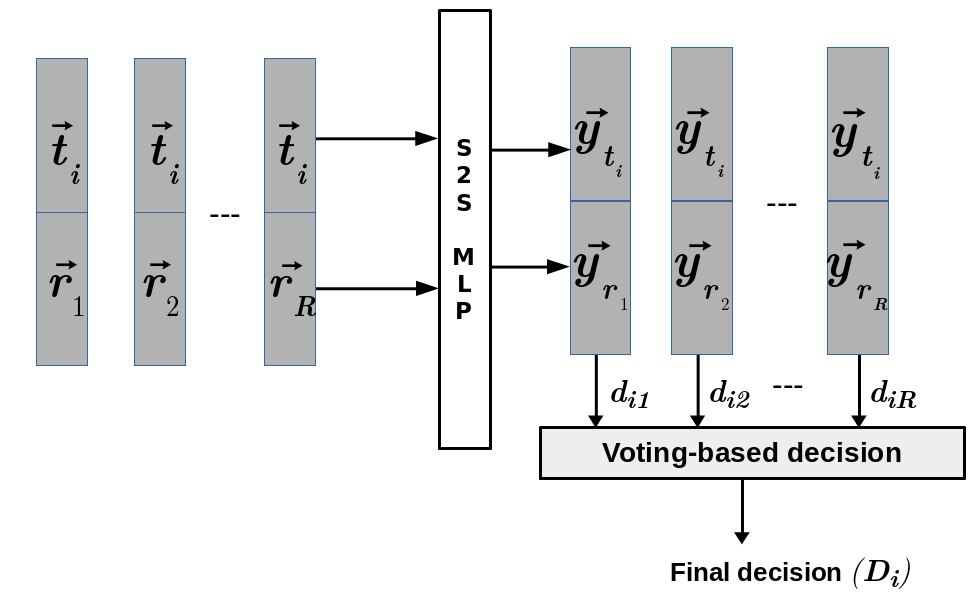}
    \caption{Testing of s$2$s-MLP.}
    \label{testing}
  \end{minipage}
\end{figure}

\section{Proposed approach}

The s$2$sL approach proposed to address low data resource problem is explained in this Section. In this work, we use MLP (modified to handle our data representation)
as the base classifier. Here, we explain the \STSL\ approach by considering two-class classification task.

\subsection{Data representation}
Consider a two-class classification task with $C = \{C_1, C_2\}$ denoting the set of class labels, and let $N_1$ and $N_2$ be the number of samples corresponding to $C_1$ 
and $C_2$, respectively.
In general, to train a classifier, the samples in the train set are provided as input-output pairs as follows.
\begin{equation}
(\vec{x}_{ij}, C_i),\quad i = 1,2;\; \textrm{and} \quad j = 1, 2, \cdots, N_i,
\end{equation}
where $\vec{x}_{ij} \in \mathbb{R}^{d\times 1}$ refers to the $d$-dimensional feature vector representing the $j^{th}$ sample corresponding to $i^{th}$ class label, and
$C_i \in C$, refers to output label of $i^{th}$ class.
In the proposed data representation format, called simultaneous two sample (s$2$s) representation, we will simultaneously consider two samples as follows.
\begin{equation}
\left ([\vec{x}_{ij}, \vec{x}_{kl}],[C_i, C_k] \right), \quad \forall i, k = 1, 2;\: j = 1, 2, \cdots, N_i\, \quad \textrm{and} \quad l=1, 2, \cdots, N_k, 
\end{equation}
where $\vec{x}_{ij}$, $\vec{x}_{kl} \in \mathbb{R}^{d\times 1}$ are the $d$-dimensional feature vectors representing the
$j^{th}$ sample in $i^{th}$ class and $l^{th}$ sample in the $k^{th}$ class, respectively; and $(C_i$, $C_k) \in C$ refers to the output labels of $i^{th}$ and $k^{th}$
class, respectively.

Hence, in s$2$s data representation, we will have an input feature vector of length $2d$ i.e., $[\vec{x}_{ij}, \vec{x}_{kl}] \in \mathbb{R}^{2d\times 1}$, and output class
labels as either $[C_1,C_1], [C_1,C_2], [C_2,C_1]$ or $[C_2,C_2]$. 
Each sample can be combined with all the samples (i.e., with ($N_1 + N_2$) samples) in the dataset. Therefore, by representing the data in the s$2$s format, the number of 
samples in the train set increases to $(N_1 + N_2)^2$ from $(N_1 + N_2)$ samples.
We hypothesize that the s$2$s format is expected to help the network not only to learn the characteristics of the two classes separately, but also the difference
and similarity in characteristics of the two classes.

\subsection{Classifier training} MLP, the most commonly used feed forward neural network, is considered as the base classifier to validate our proposed \STS\ framework.
Generally, MLPs are trained using the data format given by eq. $(1)$. But to train the MLP on our \STS\ based data representation (as in eq. $(2)$), the following modifications are made to
the MLP architecture (refer to Figure \ref{ProposedMethod}).
\begin{itemize}
 \item We have $2\times d$ units (instead of $d$ units) in the input layer to accept the two samples i.e., $\vec{x}_{ij}$ and $\vec{x}_{kl}$, simultaneously.
 \item The structure of the hidden layer in this approach is similar to that of a regular MLP. The number of hidden layers and hidden
 units can be varied depending upon the complexity of the problem. The number of units in the hidden layer is selected empirically by
 varying the number of hidden units from $2$ to twice the length of the input layer (i.e., $2$ to $4\times d$) and the unit at which the 
 highest performance is obtained are selected. In this paper, we considered only a single hidden layer. Rectified linear units (ReLU) are used for hidden layer.
 \item The output layer will consist of units equal to twice the considered number of classes in the classification task i.e, the 
 output layer will have four units for two-class classification task. The sigmoid activation function (not softmax) is used on the
 output layer units.
Unlike regular MLP, we use sigmoid activation units in the output layer, with binary
cross-entropy as the cost function, because the output labels in the proposed \STS\ based data representation will have more than one unit active at a time
(not one-hot encoded output) and this condition cannot be handled using softmax function.
\end{itemize}
As can be seen from Figure \ref{ProposedMethod}, the output layer in our proposed method has outputs $\vec{y}_i$ and $\vec{y}_k$ which corresponds to the outputs
associated with the input feature vectors $\vec{x}_{ij}$ and $\vec{x}_{kl}$, respectively. For a two-class classification problem, there will be four units in the
output layer and the possible output labels are $[0,1,0,1]$, $[0,1,1,0]$, $[1,0,0,1]$, $[1,0,1,0]$ corresponding to the class labels $[C_1,C_1], [C_1,C_2], [C_2,C_1]$ and
$[C_2,C_2]$, respectively. This architecture is referred to as s$2$s-MLP. In s$2$sL, s$2$s-MLP is trained using the s$2$s data representation format. Further, the s$2$s-MLP 
is trained using adam optimizer.

\subsection{Classifier testing}

Generally, the feature vector corresponding to the test sample is provided as input to the trained MLP in the testing phase and the
class label is decided based on the obtained output.

However, in \STSL\ method, the feature vector corresponding to the test sample should also be converted to the \STS\ data representation format to test the trained
s$2$s-MLP. 
We propose a testing approach, where the given test sample is combined with a
set of preselected reference samples, whose class label is known a priori, to generate multiple instances of the same test sample as follows.
\begin{equation}
 [\vec{t}, \vec{r}_j], \quad j = 1, 2, \cdots, R,
\end{equation}
where $\vec{t} \in \mathbb{R}^{d\times 1}$, $\vec{r}_j \in \mathbb{R}^{d\times 1}$ refer to the $d$-dimensional feature vector corresponding to the test sample
and the $j^{th}$ reference sample, respectively. $R$ refers to the considered number of reference samples.
These reference samples can be selected from any of the two classes.

For testing the s$2$s-MLP (as shown in Figure \ref{testing}), each test sample $\vec{t}_i$ (same as '$t$' in (3)) is combined with all the $R$ reference samples
($\vec{r}_1, \vec{r}_2, \cdots, \vec{r}_R$) to form $R$ different instances of the same test sample $\vec{t}_i$.
The corresponding outputs ($d_{i1}, d_{i2}, \cdots, d_{iR}$) obtained from s$2$s-MLP for the $R$ generated instances of $\vec{t}_i$ are combined by voting-based decision
approach to obtain the final decision $D_i$.
The class label that gets maximum votes is considered as the predicted output label.

\section{Experiments}

We validate the performance of the proposed \STSL\ by providing the preliminary results obtained on two different tasks namely, Speech/Music discrimination and emotion
classification.
We considered the GTZAN Music-Speech dataset [17], consisting of $120$ audio files ($60$ speech and $60$ music), for task of classifying speech and music.
Each audio file (of $2$ seconds duration) is represented using a $13$-dimensional mel-frequency cepstral coefficient (MFCC) vector, where each MFCC vector is the average of
all the frame level (frame size of $30$ msec and an overlap of $10$ msec) MFCC vectors. It is to be noted that our main intention for this task is not better feature
selection, but to demonstrate the effectiveness of our approach, in particular for low data scenarios.

The standard Berlin speech emotion database (EMO-DB) [18] consisting of $535$ utterances corresponding to $7$ different emotions is considered for the task of emotion
classification. Each utterance is represented by a $19$-dimensional feature vector obtained by using the feature selection algorithm from WEKA toolkit [19] on the
$384$-dimensional utterance level feature vector obtained using openSMILE toolkit [20]. For two class classification, we considered the two most confusing emotion pairs
i.e., (Neutral,Sad) and (Anger, Happy).
Data corresponding to Speech/Music classification ($60$ speech and $60$ music samples) and Neutral/Sad classification ($79$ neutral and $62$ sad utterances) is balanced
whereas Anger/Happy classification task has data imbalance, with anger forming the majority class ($127$ samples) and
happy forming the minority class ($71$ samples). Therefore, we show the performance of \STSL\ on both, balanced and imbalanced datasets.

All experimental results are validated using $5$-fold cross validation ($80$\% of data for training and $20$\% for testing in each fold). Further, to analyze the
effectiveness of \STSL\ in low resource scenarios, different proportions of training data, within each fold, are considered to train the system. For this analysis, we 
considered $4$ different proportions i.e., $(1/4)^{th}$, $(2/4)^{th}$, $(3/4)^{th}$ and $(4/4)^{th}$ of the training data to train the classifier. For instance, $(2/4)^{th}$
means considering only half of the original training data to train the classifier, and $(4/4)^{th}$ means considering the complete training data. $5$-fold cross validation 
is considered for all data proportions.
Accuracy (in \%) is used as a performance measure for balanced data classification tasks (i.e., Speech/Music classification and Neutral/Sad emotion classification), whereas
the more preferred $F_1$ measure [21] is used as a measure for imbalanced data classification task (i.e., Anger/Happy emotion classification).

Table \ref{table1} show the results obtained for proposed \STSL\ approach in comparison to that of MLP for the tasks of Speech/Music and 
Neutral/Sad classification, by considering different proportions of training data. The values in Table \ref{table1} are mean accuracies (in \%) obtained by $5$-fold
cross validation. It can be observed from Table \ref{table1} that for both tasks, \STSL\ method outperforms MLP, especially
at low resource conditions. s$2$sL shows an absolute improvement in accuracy of $4.4$\% and $4.1$\% over MLP for Speech/Music and Neutral/Sad classification
tasks, respectively, when $(1/4)^{th}$ of the original training data is used in experiments.

Table \ref{table2} show the results (in terms of $F_1$ values) obtained for proposed \STSL\ approach in comparison to that of MLP for Anger/Happy classification
(data imbalance problem). Here, state-of-the-art methods i.e., Eusboost [22] and MWMOTE [23] are also considered for comparison. 
It can be observed from Table \ref{table2} that the \STSL\ method outperforms MLP, and also performs better than Eusboost and MWMOTE techniques on imbalanced data
(around $3$ \% absolute improvement in $F_1$ value for s$2$sL compared to MWMOTE, when $(4/4)^{th}$ of the training data is considered).
In particular, at lower amounts of training data, \STSL\ outperforms all the other methods, illustrating its effectiveness even for low resourced data imbalance problems.
\STSL\ method shows an absolute improvement of $6$\% ($0.54 - 0.48$) in $F_1$ value over the second best ($0.48$ for MWMOTE), when only $(1/4)^{th}$ of the training data is
used.

\begin{table}[t]
  \begin{minipage}{.42\textwidth}
  \centering
    \caption{Accuracies (in \%) for balanced data classification.}
    \label{table1}
  \begin{tabular}{llllll}
    \toprule
    & &\multicolumn{4}{c}{Data proportion}\\
    \cmidrule{3-6}
    Task & & 1/4 & 2/4 & 3/4 & 4/4 \\
    \cline{3-6}
    \midrule
    Speech/ & MLP &70.8&74.6&80.1& 81.2\\
    Music&\textbf{ \STSL\ }&\textbf{75.2}&\textbf{79.3}&\textbf{82.7}&\textbf{85.1}\\
    \midrule
    Neutral/ & MLP &86.3&88.0&90.5&91.1\\
    Sad&\textbf{ \STSL\ }&\textbf{90.4}&\textbf{91.2}&\textbf{92.1}&\textbf{92.9}\\
    \bottomrule
  \end{tabular}
  \end{minipage}\hfill
  \begin{minipage}{.45\textwidth}
  \centering
  \caption{$F_1$ for imbalanced data classification. Note: EB is Eusboost and MM is MWMOTE.}
  \label{table2}
  \begin{tabular}{llllll}
    \toprule
    &&\multicolumn{4}{c}{Data proportion}\\
    \cmidrule{3-6}
    Task & & 1/4 & 2/4 & 3/4 & 4/4 \\
    \cline{3-6}
    \midrule
    & MLP &.41&.49&.53&.56\\
   Anger/ & EB &.47&.54&.59&.64\\
    Happy& MM &.48&.55&.61&.66\\
    &\textbf{ \STSL\ }&\textbf{.54}&\textbf{.60}&\textbf{.64}&\textbf{.69}\\
    \bottomrule
  \end{tabular}
  \end{minipage}\hfill
\end{table}

\section{Conclusions}
In this paper, we introduced a novel s$2$s framework to effectively learn the class discriminative characteristics, even from low data resources.
In this framework, more than one sample (here, two samples) are simultaneously considered to train the classifier. Further, this framework allows to generate multiple
instances of the same test sample, by considering preselected reference samples, to achieve a more profound decision making.
We illustrated the significance of our approach by providing the experimental results for two different tasks namely, speech/music discrimination and emotion classification.
Further, we showed that the s$2$s framework can also handle the low resourced data imbalance problem.

%

\section*{References}
\small
%

[1] Hinton, G., Deng, L., Yu, D., Dahl, G. E., Mohamed, A. R., Jaitly, N., Senior, A., Vanhoucke, V., Nguyen, P., Sainath, T. N.\ \& Kingsbury, B.\ (2012) Deep neural 
networks for acoustic modeling in speech recognition: The shared views of four research groups. {\it IEEE Signal Processing Magazine}, pp.\ 82--97.

[2] Vinyals, O., Ravuri, S. V.\ \& Povey, D.\ (2012) Revisiting recurrent neural networks for robust ASR. In {\it Proc. IEEE International Conference on Acoustics, Speech
and Signal Processing (ICASSP)}, pp.\ 4085--4088.

[3] Lu, L., Zhang, X., Cho K.\ \& Renals, S.\ (2015) A study of the recurrent neural network encoder-decoder for large vocabulary speech recognition. 
In {\it Proc. INTERSPEECH}, pp.\ 3249--3253.

[4] Zhang, Y., Pezeshki, M., Brakel, P., Zhang, S., Bengio, C.L.Y.\ \& Courville, A.\ (2017) Towards end-to-end speech recognition with
deep convolutional neural networks. {\it arXiv preprint}, arXiv:1701.02720.

[5] Han, K., Yu, D.\ \& Tashev, I.\ (2014) Speech emotion recognition using deep neural network and extreme learning machine. In {\it Proc. Interspeech}, pp.\ 223--227.

[6] Trigeorgis, G., Ringeval, F., Brueckner, R., Marchi, E., Nicolaou, M. A., Schuller, B.,\ \& Zafeiriou, S.\ (2016) Adieu features? End-to-end speech emotion
recognition using a deep convolutional recurrent network. In {\it Proc. ICASSP}, pp.\ 5200--5204).

[7] Huang, Che-Wei,\ \& Shrikanth S. Narayanan.\ (2016) Attention Assisted Discovery of Sub-Utterance Structure in Speech Emotion Recognition. In {\it Proc. INTERSPEECH},
pp.\ 1387--1391.

[8] Zhang, Z., Ringeval, F., Han, J., Deng, J., Marchi, E.\ \& Schuller, B.\ (2016) Facing realism in spontaneous emotion recognition from speech: Feature
enhancement by autoencoder with LSTM neural networks. In {\it Proc. INTERSPEECH}, pp.\ 3593--3597.

[9] Scheirer, E. D.\ \& Slaney, M.\ (2003) Multi-feature speech/music discrimination system. {\it U.S. Patent 6,570,991}.

[10] Pikrakis, A.\ \& Theodoridis S.\ (2014) Speech-music discrimination: A deep learning perspective. In {\it Proc. European Signal Processing Conference (EUSIPCO)}, 
pp.\ 616--620.

[11] Choi, K., Fazekas, G.\ \& Sandler, M.\ (2016) Automatic tagging using deep convolutional neural networks. {\it arXiv preprint}, arXiv:1606.00298.

[12] Zazo Candil, R., Sainath, T.N., Simko, G.\ \& Parada, C.\ (2016) Feature learning with raw-waveform CLDNNs for Voice Activity Detection. In {\it Proc. Interspeech},
pp.\ 3668--3672.

[13] Tian, L., Moore, JD. \ \& Lai, C.\ (2015) Emotion recognition in spontaneous and acted dialogues. In {\it Proc. International Conference on Affective Computing and 
Intelligent Interaction (ACII)}, pp.\ 698--704.

[14] Dumpala, S.H., Chakraborty, R.\ \& Kopparapu, S.K.\ (2017) k-FFNN: A priori knowledge infused feed-forward neural networks. {\it arXiv preprint}, arXiv:1704.07055.

[15] Dumpala, S.H., Chakraborty, R.\ \& Kopparapu, S.K.\ (2018) Knowledge-driven feed-forward neural network for audio affective content analysis. {\it To appear in AAAI-2018 workshop on Affective Content Analysis}.

[16] Deng, J., Zhang, Z., Marchi, E.\ \& Schuller, B.\ (2013) Sparse autoencoder-based feature transfer learning for speech emotion recognition.
In {\it Proc. Humaine Association Conference on Affective Computing and Intelligent Interaction (ACII)}, pp.\ 511--516.

[17] George Tzanetakis, Gtzan musicspeech. Availabe on-line at http://marsyas.info/download/datasets.

[18] Burkhardt, F., Paeschke, A., Rolfes, M., Sendlmeier, W.F.\ \& Weiss, B.\ (2005) A database of german emotional speech. In {\it Proc. Interspeech}, pp.\ 1517--1520.

[19] Hall, M., Frank, E., Holmes, G., Pfahringer, B., Reutemann, P.\ \& Witten, I.H.\ (2009) The WEKA data mining software: an update. {\it ACM SIGKDD explorations
newsletter}, {\bf 11}(1), pp.\ 10--18.

[20] Eyben, F., Weninger, F., Gross, F.\ \& Schuller, B.\ (2013) Recent developments in opensmile, the munich open-source multimedia feature extractor.
In {\it Proc. ACM international conference on Multimedia}, pp.\ 835--838.

[21] Maratea, A., Petrosino, A.\ \& Manzo, M.\ (2014) Adjusted f-measure and kernel scaling for imbalanced data learning. {\it Information Sciences} {\bf 257}:331--341.

[22] Galar, M., Fernandez, A., Barrenechea, E.\ \& Herrera, F.\ (2013) Eusboost: Enhancing ensembles for highly imbalanced data-sets by evolutionary undersampling.
{\it Pattern Recognition} {\bf 46}(12):3460--3471.

[23] Barua, S., Islam, M. M., Yao, X.\ \& Murase, K.\ (2014) Mwmote–majority weighted minority oversampling technique for imbalanced data set learning. {\it IEEE 
Transactions on Knowledge and Data Engineering} {\bf 26}(2):405--425.




\end{document}